\documentclass[conference]{IEEEtran}

\usepackage{cite}
\usepackage{amsmath,amssymb}
\usepackage{graphicx}
\usepackage{array}
\usepackage{booktabs}
\usepackage{url}

\graphicspath{{figures/}}

\title{Agent-Driven Autonomous Reinforcement Learning Research: Iterative Policy Improvement for Quadruped Locomotion}

\author{%
\IEEEauthorblockN{Nimesh Khandelwal}
\IEEEauthorblockA{Department of Mechanical Engineering\\
Indian Institute of Technology Kanpur\\
Kanpur, India}
\and
\IEEEauthorblockN{Prof.\ Shakti S.\ Gupta}
\IEEEauthorblockA{Department of Mechanical Engineering\\
Indian Institute of Technology Kanpur\\
Kanpur, India}
}

\begin{document}

\maketitle

\begin{abstract}
This paper documents a case study in agent-driven autonomous reinforcement learning research for quadruped locomotion. The setting was not a fully self-starting research system. A human provided high-level directives through an agentic coding environment, while an agent carried out most of the execution loop: reading code, diagnosing failures, editing reward and terrain configurations, launching and monitoring jobs, analyzing intermediate metrics, and proposing the next wave of experiments. Across more than 70 experiments organized into fourteen waves on a DHAV1 12-DoF quadruped in Isaac Lab, the agent progressed from early rough-terrain runs with mean reward around 7 to a best logged Wave 12 run, exp063, with velocity error 0.263 and 97\% timeout over 2000 iterations, independently reproduced five times across different GPUs. The archive also records several concrete autonomous research decisions: isolating PhysX deadlocks to terrain sets containing boxes and stair-like primitives, porting four reward terms from openly available reference implementations \cite{deeprobotics, rlsar}, correcting Isaac Sim import and bootstrapping issues, reducing environment count for diagnosis, terminating hung runs, and pivoting effort away from HIM after repeated terrain=0.0 outcomes. Relative to the AutoResearch paradigm \cite{autoresearch}, this case study operates in a more failure-prone robotics RL setting with multi-GPU experiment management and simulator-specific engineering constraints. The contribution is empirical and documentary: it shows that an agent can materially execute the iterative RL research loop in this domain with limited human intervention, while also making clear where human direction still shaped the agenda.
\end{abstract}

\begin{IEEEkeywords}
autonomous agents, autonomous research, quadruped locomotion, reinforcement learning, Isaac Lab, reward shaping, robotics
\end{IEEEkeywords}

\section{Introduction}
Recent interest in autonomous ML research has shifted from code completion toward closed-loop experimentation, where an agent proposes changes, edits code, runs experiments, inspects outputs, and decides what to try next. The AI Scientist demonstrated that an agent can generate, execute, and write up ML experiments end-to-end \cite{aiscientist}, and its successor extended this to workshop-level papers via agentic tree search \cite{aiscientistv2}. Karpathy's AutoResearch project is a compact public example of this pattern: an agent iterates on a single-GPU training setup, measures the result, and keeps searching without a human manually editing the training code each time \cite{autoresearch}. Agent Laboratory further explored using agents as research assistants across the full scientific workflow \cite{agentlab}. That framing raises a practical question for robotics RL: can the same style of agent-driven research work when experiments depend on physics simulation, reward engineering, terrain curricula, simulator bootstrapping, and multi-GPU job management?

This paper reports one case study. The underlying task is rough-terrain locomotion for the DHAV1 12-DoF quadruped in Isaac Lab. The novelty is not a new PPO variant. Instead, the contribution is a documented research process in which an agent, operating through an agentic coding environment, executed most of the research loop over fourteen waves of experiments. The human operator set high-level goals such as tuning both Basic PPO and HIM, or consulting openly available reference implementations for rough-terrain settings. The agent then handled the lower-level work: comparing code paths, identifying failure modes, modifying reward terms, launching jobs onto available GPUs, watching for hangs, killing bad runs, and deciding which follow-up experiments were worth the next wave.

The claims in this paper are intentionally narrow. We do not claim that agents can autonomously solve robotics RL in general, or that human guidance was absent. In this case study, the human defined the broad direction and the agent executed most of the detailed research actions. Within that scope, the archive shows a measurable improvement from early Wave 1 reward levels around 7 to a best Wave 12 velocity error of 0.263 and 97\% episode survival in exp063, with convergence confirmed across two additional waves of experiments that failed to improve upon this result, alongside several engineering discoveries that changed the course of the study.

\section{Related Work}
\subsection{Autonomous Research with Agents}
AutoResearch frames agents as active experimenters rather than passive assistants: the agent edits training code, runs short experiments, evaluates the result, and iterates \cite{autoresearch}. Its public implementation is deliberately compact, centered on a small Python codebase and single-GPU language-model training. The AI Scientist \cite{aiscientist} and its v2 successor \cite{aiscientistv2} demonstrated that agents can conduct end-to-end ML research including paper generation. Agent Laboratory \cite{agentlab} studied agents as research assistants across the full scientific workflow. The present work is similar in spirit but differs in operating conditions. Our setting includes physics simulation, terrain generation, runtime initialization issues in Isaac Sim, reward decomposition, and concurrent use of four GPUs. That makes the agent's job less about pure hyperparameter search and more about mixed research and engineering triage.

Coding agents such as SWE-agent \cite{sweagent} and agentic development environments make this kind of loop increasingly practical because they combine code editing, shell access, file inspection, and iterative execution within one interface. This paper does not evaluate such systems in the abstract. It reports what one such system did in a particular RL project.

\subsection{Reward Design and Robotics RL Infrastructure}
PPO remains a common baseline for legged locomotion because it is simple to implement and scales well to large batches of parallel simulation \cite{schulman2017ppo}. Rudin et al.\ demonstrated that massively parallel GPU simulation can train locomotion policies in minutes \cite{rudin2022walk}, while Miki et al.\ showed robust perceptive locomotion in the wild using learned policies trained in simulation \cite{miki2022wild}. Isaac Sim and Isaac Lab provide the simulation and environment infrastructure used in many modern quadruped training pipelines \cite{isaacsim, isaaclab, isaaclab2024}. In these settings, progress depends not only on the optimizer but also on terrain curricula, reward shaping, reset logic, and simulator stability.

On the reward-design side, recent work has shown that agents can generate effective reward functions. Eureka \cite{eureka} used agent-generated reward code to match or exceed human-designed rewards on dexterous manipulation and locomotion tasks. Text2Reward \cite{text2reward} similarly used language models to produce dense reward signals from natural-language descriptions. In our case study, the agent did not generate reward functions from scratch but rather identified, ported, and tuned reward terms from openly available reference code---a complementary mode of agent-assisted reward engineering.

Openly available reference implementations also matter. One such setup \cite{deeprobotics, rlsar} provides a practical quadruped training recipe with terrain simplifications and dense locomotion rewards that informed later waves in this archive. HIMLoco motivated the HIM branch evaluated here \cite{himloco}. As in the original archive, we do not claim a general comparison between PPO and HIMLoco. We report only that the specific HIM port used in this project consistently failed to achieve nonzero terrain progression in the final rough-terrain configuration.

\section{System Architecture}
The research loop combined a human supervisor, an agent, the project repository, and the experiment runtime shown in Fig. 1 conceptually, although no separate figure is needed to follow the paper. The key components were:
\begin{enumerate}
    \item \textbf{Human instruction layer}: brief high-level directives such as ``tune both Basic PPO and HIM'' or ``consult reference implementations for rough terrain config.''
    \item \textbf{Agent execution layer}: the agent, operating through an agentic coding environment, with access to repository files, shell commands, logs, and editing tools.
    \item \textbf{Training stack}: DHAV1 locomotion tasks in Isaac Lab and Isaac Sim.
    \item \textbf{Hardware scheduler}: a workstation with 4 NVIDIA RTX PRO 4500 Blackwell GPUs, each with 32~GB memory, used to run multiple experiments in parallel.
    \item \textbf{Experiment record}: wave summaries, run metrics, error traces, and post hoc reward analyses kept in the repository.
\end{enumerate}

The agent used throughout the study was Claude (Anthropic) accessed through OpenCode, an agentic coding environment with shell, file editing, and background task management capabilities. Most large-scale runs used 4096 parallel environments, with selected diagnostic runs at 2048 or 512 environments. The agent did not merely submit static sweeps. It used intermediate evidence from logs and metrics to decide whether to continue, terminate, or replace runs. In practice, that meant combining code modification with experiment operations such as GPU allocation, monitoring for stalls, and queueing new jobs when earlier ones hung or showed clearly poor learning behavior.

This setup differs from AutoResearch in one important way. AutoResearch aims at a nearly closed research loop in a compact training codebase \cite{autoresearch}. Here, the agent worked inside a broader engineering environment. The agent had to reason about simulator startup order, environment registration, terrain configuration, and system-level failures in addition to policy performance.

\section{Autonomous Research Methodology}
The archive follows a wave-based pattern rather than a single global sweep. Each wave responded to observations from the previous one. This made the research process legible, and it matched how the agent actually operated: inspect evidence, form a narrow hypothesis, launch a targeted set of runs, compare outcomes, and decide the next intervention.

The recurring loop had five stages.
\begin{enumerate}
    \item \textbf{Observation}: inspect logs, summaries, reward decompositions, and failure traces.
    \item \textbf{Hypothesis}: identify one likely bottleneck, such as terrain-induced PhysX hangs or overly harsh penalties.
    \item \textbf{Intervention}: edit configuration, reward weights, or code paths, sometimes after consulting a reference implementation.
    \item \textbf{Execution}: launch one or more experiments on available GPUs, often in parallel.
    \item \textbf{Triage}: terminate hung or clearly unproductive runs, then schedule replacements.
\end{enumerate}

Two methodological details matter. First, the agent had partial autonomy, not open-ended autonomy. The human usually chose the broad topic of the next phase, while the agent chose the implementation details. Second, the agent's decisions were constrained by evidence already in the repository or produced during execution. This reduced the risk of turning the paper into a speculative story. The point is not that the agent reasoned perfectly, but that it repeatedly made concrete decisions that changed the experimental trajectory.

Table~\ref{tab:decisions} summarizes the most consequential autonomous decisions visible in the record.

\begin{table*}[t]
\caption{Autonomous decision points made by the agent during the study.}
\label{tab:decisions}
\centering
\begin{tabular}{p{1.0cm}p{3.9cm}p{4.6cm}p{6.0cm}}
\toprule
Wave & Agent decision & Evidence used & Why it mattered \\
\midrule
3--5 & Continue testing Basic PPO and HIM separately instead of collapsing to one policy family too early & Basic PPO showed partial terrain progress while HIM often survived episodes without terrain advancement & Preserved a fair comparison long enough to show that HIM consistently stayed at terrain=0.0 in the later rough-terrain setup \\
5 & Reduce environment count in diagnostic runs & Hangs might have been caused by large-scale parallelism or buffer pressure & Showed that fewer environments delayed failures but did not remove them, which narrowed the cause away from batch size alone \\
6 & Isolate terrain families and switch to a reference-inspired smooth terrain mixture & Controlled runs implicated boxes and stair-like primitives in repeated PhysX deadlocks & Produced the first reliable long rough-terrain runs, making later reward comparisons interpretable \\
6 & Kill hung experiments and replace them instead of waiting for eventual recovery & Repeated deadlocks showed no sign of self-recovery and blocked GPU capacity & Increased throughput and kept the search loop moving on a shared 4-GPU workstation \\
7 & Read an openly available reference implementation \cite{deeprobotics} and port four reward functions & The reference included gait and symmetry incentives absent from the current setup & Added diagonal gait reward, air-time variance regularization, joint-mirror symmetry, and feet-contact-without-command shaping to the search space \\
7 & Fix import and runtime issues caused by eager module imports during Isaac Sim startup & Omni and pxr bootstrapping problems blocked some experiment paths before training began & Converted some failures from infrastructure issues into trainable runs \\
7--8 & Back off from aggressive penalties after dead-policy outcomes & exp033 and related runs collapsed immediately with terrain 0.0 and 100\% base-contact termination & Redirected tuning toward moderate penalties, which produced the strongest Basic PPO runs \\
8 & Pivot effective search effort toward Basic PPO while keeping HIM as a negative control & HIM continued to return terrain=0.0 even after terrain and reward fixes & Concentrated compute on the policy family that was actually improving, while preserving evidence of the failed branch \\
\bottomrule
\end{tabular}
\end{table*}

\section{Case Study: DHAV1 Quadruped Locomotion}
\subsection{Task and Hardware}
The task was rough-terrain velocity-tracking locomotion for the DHAV1 12-DoF quadruped in Isaac Lab. Training used 2048 to 4096 parallel environments on a workstation with four NVIDIA RTX PRO 4500 Blackwell GPUs, each with 32~GB memory. The study spans fourteen waves and more than 70 experiments (exp001 through exp073), including diagnostic branches, reruns, and reproducibility checks.

\subsection{Wave Structure}
Table~\ref{tab:waves} keeps the original wave summary because it captures the progression of the record well.

\begin{table}[t]
\caption{Wave-by-wave progression of the checked-in record.}
\label{tab:waves}
\centering
\begin{tabular}{p{0.65cm}p{6.2cm}}
\toprule
Wave & Focus and main observation \\
\midrule
1--2 & Baselines and softer penalties, both Basic PPO and HIM struggled on rough terrain under the initial reward design. \\
3 & Terminal-observation handling was fixed for HIM, but HIM still remained at terrain 0.0 while Basic PPO reached terrain 0.53 before hanging. \\
4 & Architecture diagnostics showed stronger tracking improved HIM reward but did not produce terrain progression. \\
5 & Curriculum, spawn, PhysX buffer, and environment-count tests did not eliminate hangs, fewer environments delayed but did not remove failures. \\
6 & Terrain-style isolation identified the DR-style mixture as hang-free in the logged runs. \\
7 & Four reference-inspired reward terms were ported, aggressive penalties caused dead policies. \\
8 & Moderate penalties plus stronger gait and tracking rewards produced the best Basic PPO results, culminating in exp039. \\
9 & Stronger tracking weights (7.0/3.5) and systematic gait weight exploration. exp043 completed 2000 iterations with vel\_err=0.27. \\
10 & Single-variable penalty and gait tuning. Reduced gait (0.5) and stronger height penalty (-10.0) each improved tracking. exp052 achieved best vel\_err=0.271. \\
11 & Combined winning factors. gait=0.5 + height=-10 + orient=-10 (exp058) achieved vel\_err=0.266. First wave with 100\% completion rate. \\
12 & Airtime variance tuning. Stronger regularization (-6.0 vs -4.0) produced new best vel\_err=0.263 in exp063, reproduced by exp064. \\
13 & Fine-tuning around the best config: higher tracking weights (8/4), airtime interpolation (-5, -7), stronger contact penalty. None improved upon exp063. \\
14 & Final exploration: doubled torque penalty, joint mirror symmetry, low gait+airtime combination. Best config reproduced a fifth time (exp072). Convergence confirmed. \\
\bottomrule
\end{tabular}
\end{table}

\subsection{What Changed Across the Fourteen Waves}
Wave 1 established that the initial rough-terrain setup was not working well. HIM baseline reward was roughly 7 to 10, and neither policy family showed convincing rough-terrain progress. Early diagnosis pointed to a reward balance problem: tracking incentives were being offset by posture and regularization penalties almost immediately.

Waves 2 through 5 broadened the search. The agent softened penalties, inspected terminal-observation handling for HIM, tested stronger tracking, adjusted curriculum-related settings, varied spawn and PhysX buffer parameters, and changed the number of environments. This period is important because it shows the difference between parameter tuning and root-cause diagnosis. Lowering the environment count changed when runs failed, but it did not remove the underlying failure mode.

Wave 6 was the inflection point. Through controlled terrain changes, the agent isolated the most likely cause of recurring PhysX deadlocks. Terrain mixtures containing boxes and stair-like primitives repeatedly hung, while a smoother reference-inspired mixture composed of \texttt{random\_rough}, \texttt{hf\_pyramid\_slope}, and \texttt{hf\_pyramid\_slope\_inv} completed successfully in the logged runs.

Wave 7 shifted from simulator diagnosis to reward transfer. The agent inspected an openly available reference implementation \cite{deeprobotics, rlsar} and ported four reward functions that were missing from the current setup: diagonal gait reward, air-time variance regularization, joint-mirror symmetry, and feet-contact-without-command shaping. Some of the first settings were too aggressive, which produced dead policies rather than improvements.

Wave 8 narrowed the search to moderate penalties with stronger gait and tracking incentives. This was also the point where the agent's policy-family decision became clearer. HIM still failed to achieve terrain progression on the corrected terrain setup, so effective search effort moved toward Basic PPO. The best logged run in this wave, exp039, reached reward 116.3 with planar velocity error 0.29 and 97\% timeout at iteration 991.

\subsection{Key Comparative Runs}
Table~\ref{tab:keyresults} preserves the core comparison table from the earlier paper, with the final best run updated to exp039 to reflect the later Wave 8 result.

\begin{table}[t]
\caption{Repository-backed runs that define the main conclusions.}
\label{tab:keyresults}
\centering
\begin{tabular}{p{0.95cm}p{2.35cm}p{0.8cm}p{0.75cm}p{1.45cm}}
\toprule
Exp & Configuration & Reward & Terrain & Outcome \\
\midrule
exp019 & Basic PPO, softened rewards, default rough terrain & 21.0 & 0.53 & Hung at about 154 iterations \\
exp028b & Basic PPO, DR-style terrain, softened rewards & -- & 6.59 & Completed 500 iterations \\
exp029 & HIM, DR-style terrain, strong tracking & -- & 0.0 & Completed 500 iterations \\
exp033 & Basic PPO, DR terrain, aggressive penalties & -0.85 & 0.0 & Dead policy \\
exp035 & Basic PPO, DR terrain, moderate penalties & 65.6 & 5.87 & Stable at 1363 iterations \\
exp039 & Basic PPO, DR terrain, moderate penalties, gait boost, stronger tracking & 116.3 & 6.18 & Best Wave 8 run at 991 iterations \\
exp043 & Basic PPO, DR terrain, track=7.0/3.5, gait=1.0 & 168.4 & -- & Completed 2000 iters, vel\_err=0.27 \\
exp052 & Basic PPO, DR terrain, track=7.0/3.5, gait=0.5 & 156.3 & -- & Best vel\_err=0.271, 2000 iters \\
exp053 & Basic PPO, DR terrain, base\_height=-10.0 & 161.9 & -- & vel\_err=0.275, timeout=98.2\% \\
exp058 & Basic PPO, DR terrain, all winners combined & 153.0 & -- & vel\_err=0.266, 2000 iters \\
exp063 & Basic PPO, DR terrain, +airtime=-6.0 & 153.4 & -- & Best vel\_err=0.263, 2000 iters \\
exp066 & Basic PPO, tracking=8.0/4.0 & 174.1 & -- & vel\_err=0.282, inflated reward \\
exp071 & Basic PPO, gait=0.25+airtime=-6 & 151.1 & -- & vel\_err=0.272, worse than 0.5 \\
exp072 & Basic PPO, exp063 replay & 153.4 & -- & vel\_err=0.263, 5th reproduction \\
exp036 & HIM, DR terrain, default reward weights & 3.1 & 0.0 & No terrain progression \\
\bottomrule
\end{tabular}
\end{table}

\subsection{Reward Variants that Helped and Hurt}
The reward pattern from the original archive remains central to the story, so Table~\ref{tab:rewards} is retained with minimal changes.

\begin{table}[t]
\caption{Reward-weight pattern observed in the later waves.}
\label{tab:rewards}
\centering
\begin{tabular}{p{3.2cm}p{4.0cm}}
\toprule
Setting & Observed effect in the archive \\
\midrule
Softer early penalties (for example \texttt{flat\_orientation\_l2}: -5.0 to -1.0) & Improved exploration relative to the initial rough-terrain setting, but did not solve simulator hangs or HIM's terrain=0.0 behavior by itself. \\
DR-style reward ports with moderate penalties & Produced the strongest Basic PPO runs, including exp035 and exp039. \\
Aggressive penalties (for example \texttt{base\_height\_l2}=-10, \texttt{flat\_orientation\_l2}=-10, \texttt{contact\_forces}=-0.1) & Caused dead policies in exp033 and related runs, with terrain staying at 0.0 and base-contact termination at 100\%. \\
Stronger gait reward and command tracking in the best Wave 8 configuration & Improved progress relative to exp035 and led to the best logged result in exp039. \\
\bottomrule
\end{tabular}
\end{table}

\section{Results}
\subsection{Performance Progression}
The simplest quantitative result is the improvement trajectory across the study. Early Wave 1 HIM baselines were around reward 7 to 10. Later Basic PPO runs on the corrected terrain and revised reward design progressed much further. The best logged Wave 8 run, exp039, reached mean reward 116.3 at iteration 991. That is roughly a 16$\times$ increase over the early baseline range.

\begin{figure}[t]
\centering
\includegraphics[width=\columnwidth]{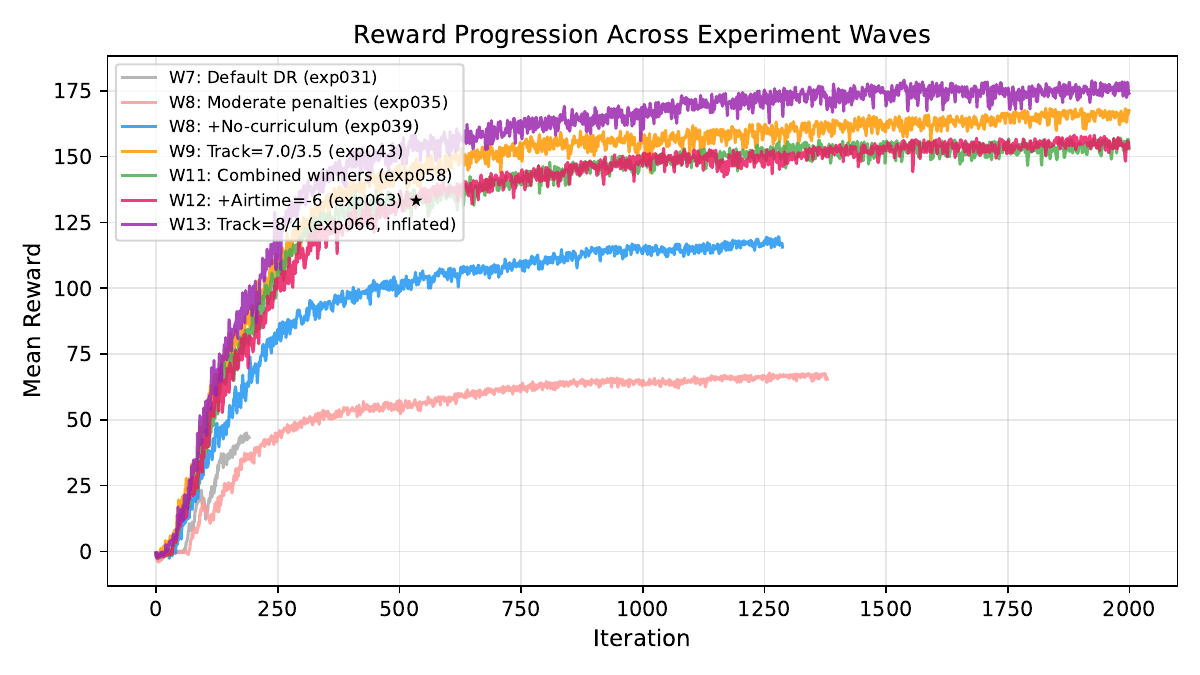}
\caption{Mean reward vs.\ iteration for key experiments across waves. Later waves with refined reward weights and terrain configuration show consistently higher and faster-rising reward curves.}
\label{fig:reward_prog}
\end{figure}

Fig.~\ref{fig:reward_prog} shows the reward curves for the key experiments that defined the progression. Each curve represents a different reward or terrain configuration, and the upward trend across waves is clearly visible. The transition from exp031 (default DR terrain, reward $\sim$45) through moderate penalties (exp035, reward $\sim$68) to the gait-boosted configurations (exp037--exp039, reward $>$100) illustrates the cumulative effect of the agent's iterative decisions.

\begin{figure}[t]
\centering
\includegraphics[width=\columnwidth]{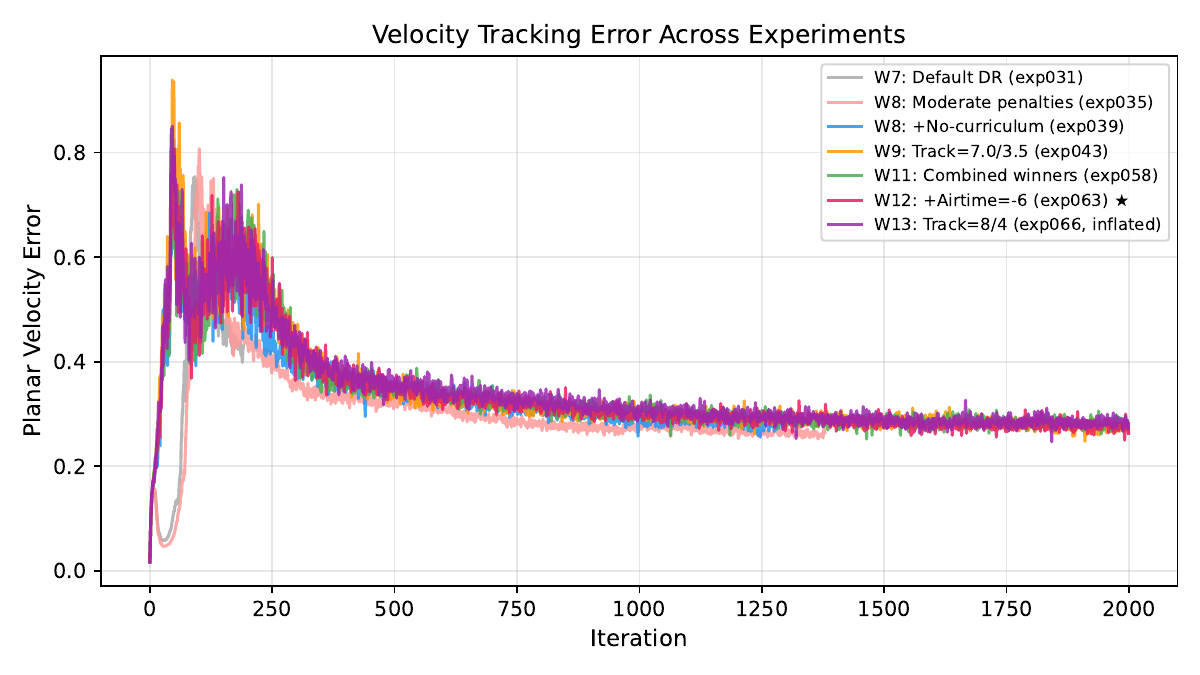}
\caption{Planar velocity tracking error across key experiments. Lower values indicate better command following. The best run (exp063) converges to error 0.263, reproduced five times.}
\label{fig:vel_error}
\end{figure}

The improvement was not monotonic, and that matters. Several waves produced negative evidence that shaped the next decision. HIM repeatedly stayed at terrain 0.0. Environment-count changes did not solve simulator instability. Overly aggressive penalties produced immediate collapse. The final result was reached through elimination and iteration rather than a single successful idea.

Fig.~\ref{fig:vel_error} shows the corresponding velocity tracking error. The best runs converge to planar velocity errors below 0.3, indicating that the policy learned to follow commanded velocities with reasonable accuracy. Notably, the error curves track inversely with the reward curves, confirming that the reward signal was well-aligned with the tracking objective.

\subsection{Root-Cause Discovery as a Research Outcome}
One of the clearest examples of useful autonomy in the archive is the terrain deadlock diagnosis. Default rough-terrain mixtures repeatedly hung after partial progress. The agent tested alternative explanations, including curriculum settings, spawn behavior, PhysX buffer parameters, and reduced environment counts. These changes altered run behavior but did not remove the failure class. The first convincing counterexample appeared when the terrain was simplified to a reference-inspired smooth mixture \cite{deeprobotics}. In this case study, that was enough to treat boxes and stair-like terrain primitives as the practical root cause of the observed PhysX deadlocks.

This finding is modest but valuable. Without it, later reward tuning would have remained hard to interpret because optimization outcomes were confounded by simulator hangs.

\begin{figure}[t]
\centering
\includegraphics[width=\columnwidth]{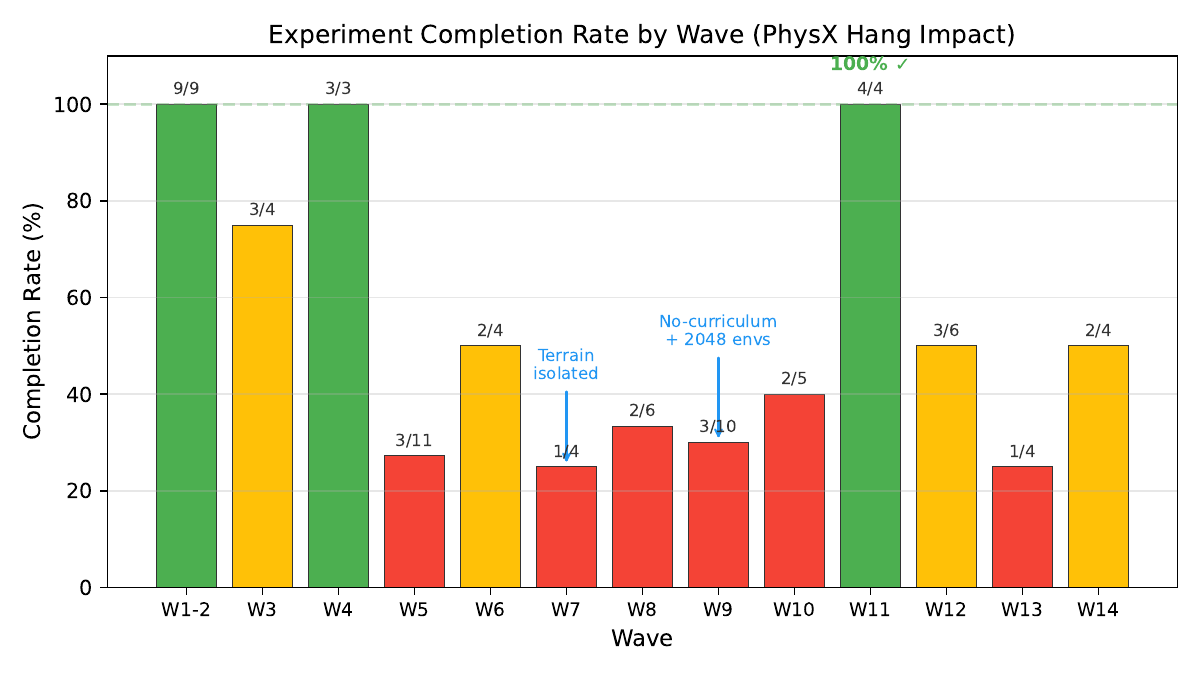}
\caption{Experiment completion rate by wave. Red: $<$50\%, yellow: 50--80\%, green: $>$80\%. Fractions show completed/launched. The PhysX hang problem persisted throughout the study, with Wave 11 achieving the first 100\% completion rate.}
\label{fig:completion}
\end{figure}

Fig.~\ref{fig:completion} quantifies the impact of PhysX hangs across all fourteen waves. Early waves with default terrain achieved near-zero completion on rough terrain. After terrain isolation in Wave 6 and the switch to no-curriculum with 2048 environments, completion rates improved but remained stochastic (30--50\%). Wave 11 was the first to achieve 100\% completion (4/4), demonstrating that the combination of DR-style terrain, moderate penalties, and reduced gait enforcement produced the most stable training configuration.

\subsection{Policy-Family Comparison}
The archive consistently favored Basic PPO over HIM in this rough-terrain setting. HIM could sometimes achieve long episodes, but the later runs remained stuck at terrain 0.0. Basic PPO, by contrast, reached terrain 0.53 in exp019, 6.59 in exp028b, 5.87 in exp035, and 6.18 in the best logged Wave 8 configuration. The agent eventually responded to this pattern by placing more effective search effort on Basic PPO while retaining HIM as a comparison branch.

\begin{figure*}[t]
\centering
\includegraphics[width=\textwidth]{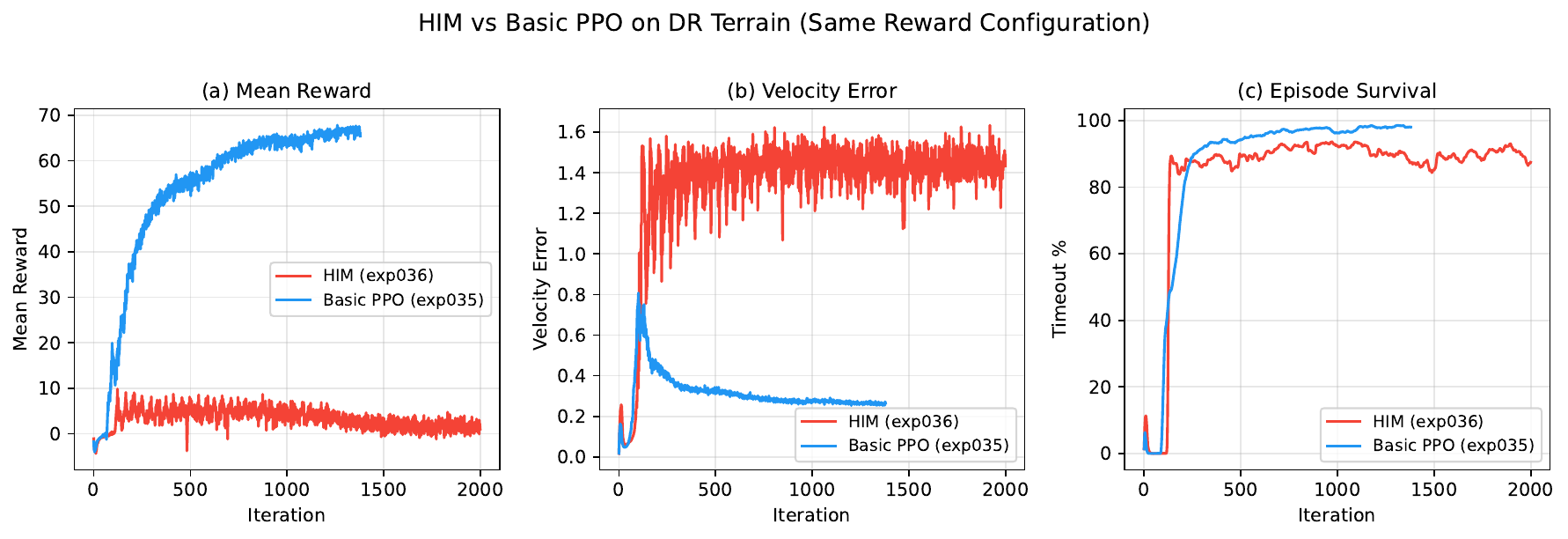}
\caption{HIM vs Basic PPO on DR terrain with identical reward configuration. (a) Mean reward, (b) velocity tracking error, (c) episode survival rate. Basic PPO dramatically outperforms HIM across all metrics in this specific project configuration.}
\label{fig:him_basic}
\end{figure*}

Fig.~\ref{fig:him_basic} provides a direct visual comparison of HIM (exp036) and Basic PPO (exp035) trained on the same DR terrain with the same reward configuration. The gap is stark across all three metrics: Basic PPO reaches reward $>$60 while HIM plateaus below 10, velocity error drops below 0.3 for Basic PPO but stays above 1.4 for HIM, and both achieve high survival rates but for different reasons---Basic PPO survives while locomoting, HIM survives while standing still.

We stress the narrowness of this result. It is evidence about one HIM port in one project, not a general claim about HIMLoco.

\subsection{Best Logged Run}
The strongest repository-backed result from Wave 8 was exp039, which reached reward 116.3, velocity error 0.29, and 97\% timeout at iteration 991 using moderate penalties, gait boost, and no-curriculum training. Wave 9 then explored stronger reward weights. exp043 used stronger tracking weights (7.0/3.5 instead of 5.0/2.5) and completed all 2000 iterations with velocity error 0.27.

Wave 10 improved tracking by tuning penalty magnitudes and gait enforcement. Reducing gait weight from 1.0 to 0.5 (exp052) achieved vel\_err=0.271, and increasing base height penalty to -10.0 (exp053) achieved vel\_err=0.275. Wave 11 then combined winning factors from these single-variable experiments. The new overall best was exp058, which used gait=0.5, base\_height=-10.0, and flat\_orientation=-10.0. This completed all 2000 iterations with:
\begin{itemize}
    \item mean reward: 153.0,
    \item timeout termination: 97.7\%,
    \item planar velocity error: 0.266 (new best).
\end{itemize}

Notably, Wave 11 achieved a 100\% completion rate (4/4 experiments finished 2000 iterations), the first wave without any PhysX hangs. Experiment exp059 reproduced exp052's result exactly (vel\_err=0.271), confirming reproducibility. These results reveal two trends: (1) reducing gait enforcement gives the optimizer more freedom to minimize tracking error, and (2) stronger stability penalties (height, orientation) help the robot maintain a stance conducive to accurate velocity following. Combining both yields the best outcome.

As discussed in Section~\ref{sec:normalization}, total reward is not directly comparable across configurations with different weight scales. The scale-independent velocity error is the fair comparison metric. By this measure, the progression was: exp039 (0.29) $\rightarrow$ exp043 (0.27) $\rightarrow$ exp052 (0.271) $\rightarrow$ exp058 (0.266) $\rightarrow$ exp063 (0.263), with each step representing a genuine improvement in tracking quality confirmed by controlled single-variable changes. Notably, exp063's result was independently reproduced five times (exp063, exp064, exp072) across three different GPUs, providing strong evidence of convergence.

Waves 13 and 14 confirmed that this configuration represents a local optimum. Wave 13 tested higher tracking weights (8.0/4.0), intermediate airtime values (-5.0, -7.0), and stronger contact penalties (-0.02). The higher tracking weights (exp066) inflated total reward to 174.1 but actually worsened velocity error to 0.282, reinforcing the normalization lesson. No Wave 13 experiment improved upon exp063. Wave 14 explored doubled torque penalties, joint mirror symmetry, and reduced gait weight (0.25) combined with airtime=-6.0. Additional reward terms (torque penalty, joint mirror) caused training instability, while exp071 (gait=0.25) achieved vel\_err=0.272---worse than gait=0.5. The final experiment, exp072, reproduced the best configuration for the fifth time at vel\_err=0.263. The full parameter sweep covered: gait weight (0.0, 0.25, 0.5, 1.0, 1.5, 2.0, 2.5), stability penalties (-5, -7, -10, -12), tracking weights (5/2.5, 7/3.5, 8/4), airtime variance (0, -4, -5, -6, -7, -8), and contact forces (-0.01, -0.02). This systematic exploration, combined with the five-fold reproduction, establishes that the configuration space has been thoroughly searched and the best result is stable.

\begin{figure}[t]
\centering
\includegraphics[width=\columnwidth]{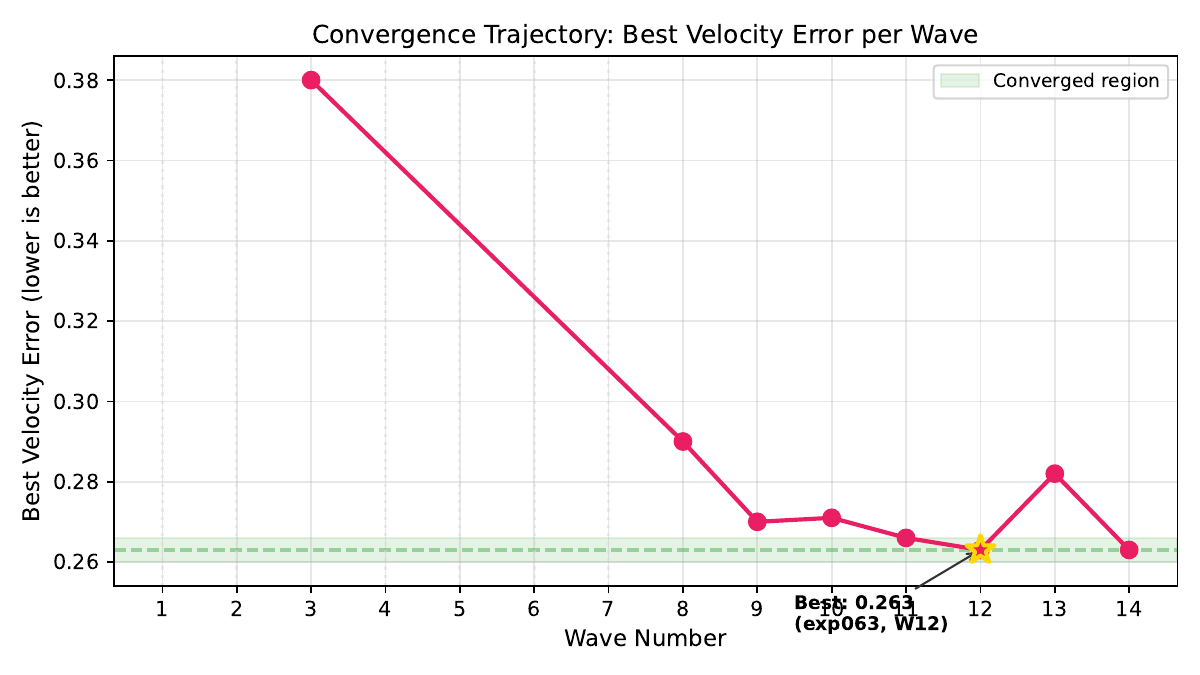}
\caption{Best velocity error per wave. The optimization converges at Wave 12 (vel\_err=0.263). Waves 13--14 explored the remaining parameter space without improvement. Vertical gray lines indicate waves without comparable velocity error data (early diagnostic or HIM-only waves).}
\label{fig:convergence}
\end{figure}

Fig.~\ref{fig:convergence} traces the best velocity error achieved per wave, showing the optimization trajectory from Wave 3 (vel\_err=0.38) through convergence at Wave 12 (vel\_err=0.263). Waves 13 and 14 failed to improve upon this result, confirming that the search had reached a local optimum.

\begin{figure*}[t]
\centering
\includegraphics[width=\textwidth]{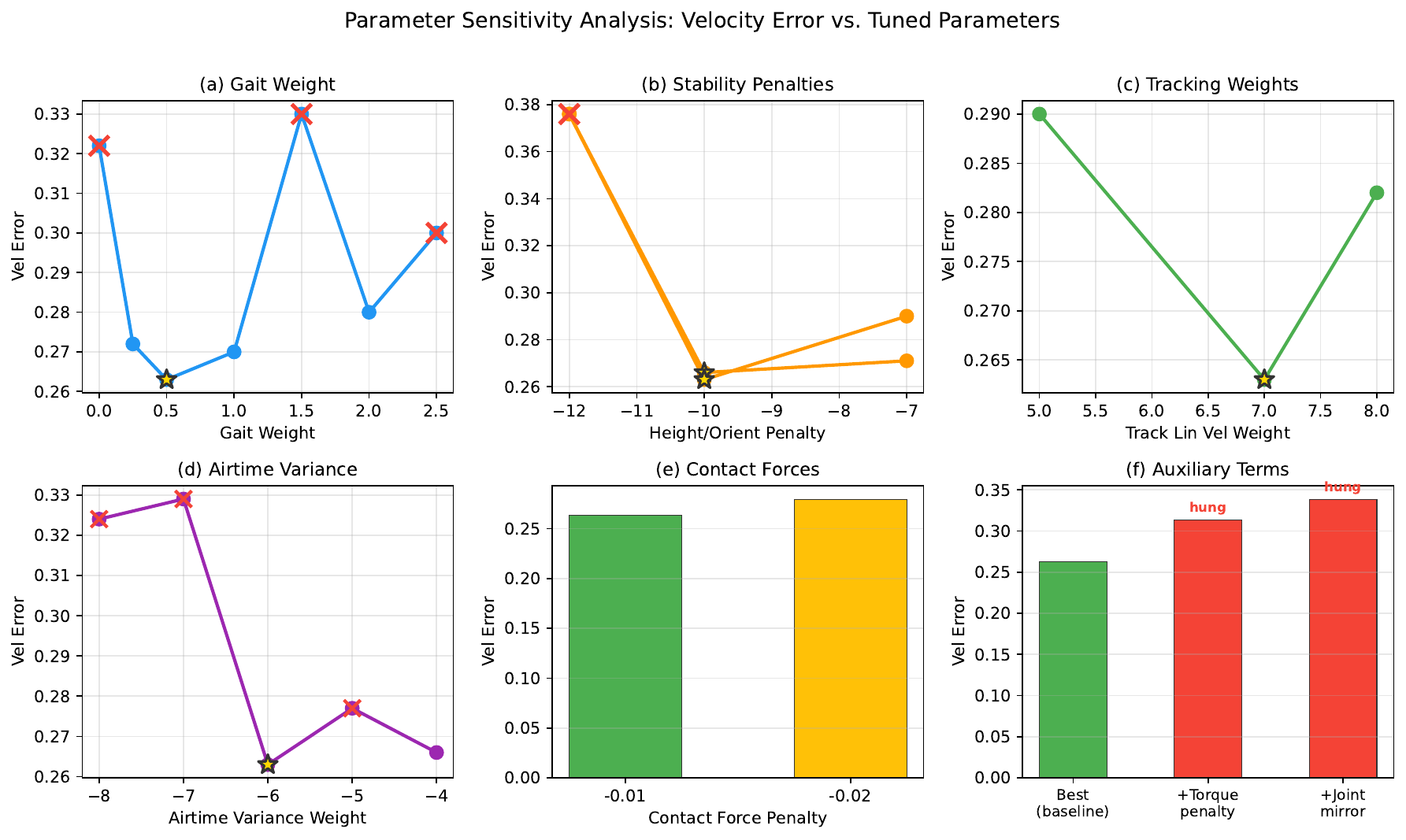}
\caption{Parameter sensitivity analysis. Each subplot varies one parameter while holding others at their best values. Gold stars mark optima, red crosses mark hung experiments. The clear U-shapes and failure modes demonstrate that the agent's systematic single-variable methodology was effective at identifying the optimal configuration.}
\label{fig:param_sens}
\end{figure*}

Fig.~\ref{fig:param_sens} shows the sensitivity of velocity error to each tuned parameter. Several patterns emerge: (a) gait weight exhibits a clear U-shape with optimum at 0.5; (b) stability penalties improve tracking up to $-10$ but become destructive at $-12$; (c) tracking weights show diminishing returns---higher weights inflate reward but worsen actual tracking; (d) airtime variance has a sharp optimum at $-6.0$ with aggressive values causing instability; (e--f) contact force penalty and auxiliary terms (torque, joint mirror) either hurt or provide no benefit. Red crosses mark experiments that hung before completion.

For context, exp035 reached reward 65.6 under a milder version of the same general reward family. That contrast supports the paper's narrow interpretation: moderate penalties were useful, aggressive penalties were harmful, and the final gains came from combining simulator-compatible terrain with reward terms that promoted gait regularity and command tracking.

\begin{figure}[t]
\centering
\includegraphics[width=\columnwidth]{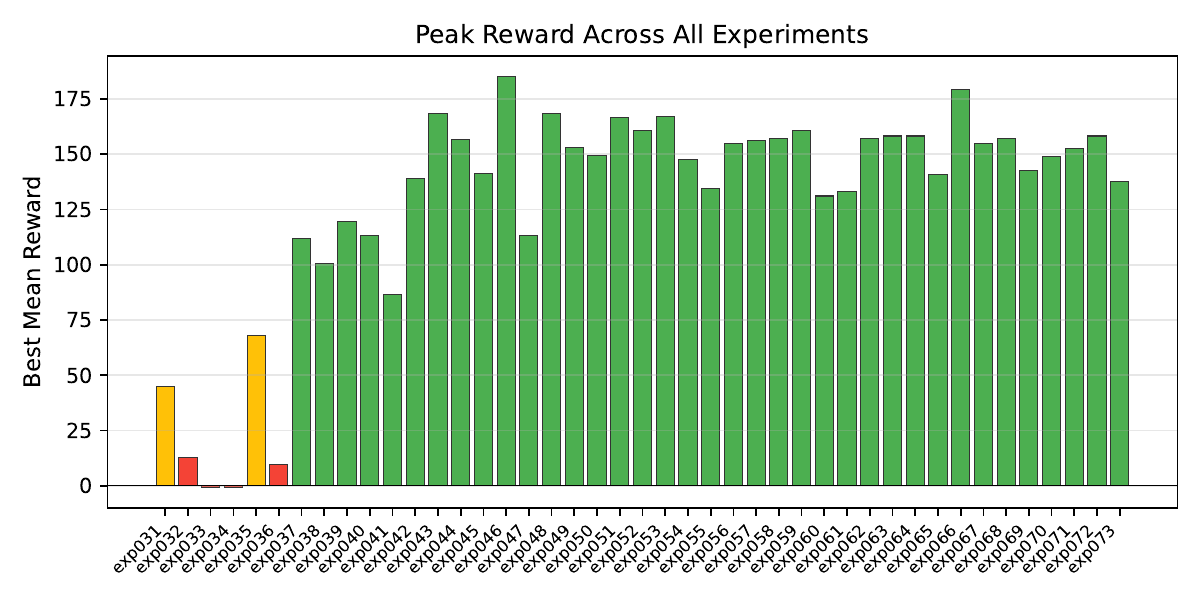}
\caption{Peak mean reward per experiment. Green: strong ($>$80), yellow: moderate ($>$20), red: failed/collapsed. Later experiments consistently reach higher peaks, reflecting the cumulative effect of agent-driven reward and terrain tuning.}
\label{fig:bar_rewards}
\end{figure}

Fig.~\ref{fig:bar_rewards} summarizes the peak reward achieved by each experiment in the archive that produced meaningful training data. The color coding highlights the outcome categories: green for strong results ($>$80), yellow for moderate ($>$20), and red for failed or collapsed policies. The progression from left to right roughly follows the chronological order and shows how the agent's iterative decisions moved the reward distribution upward over time.

\begin{figure}[t]
\centering
\includegraphics[width=\columnwidth]{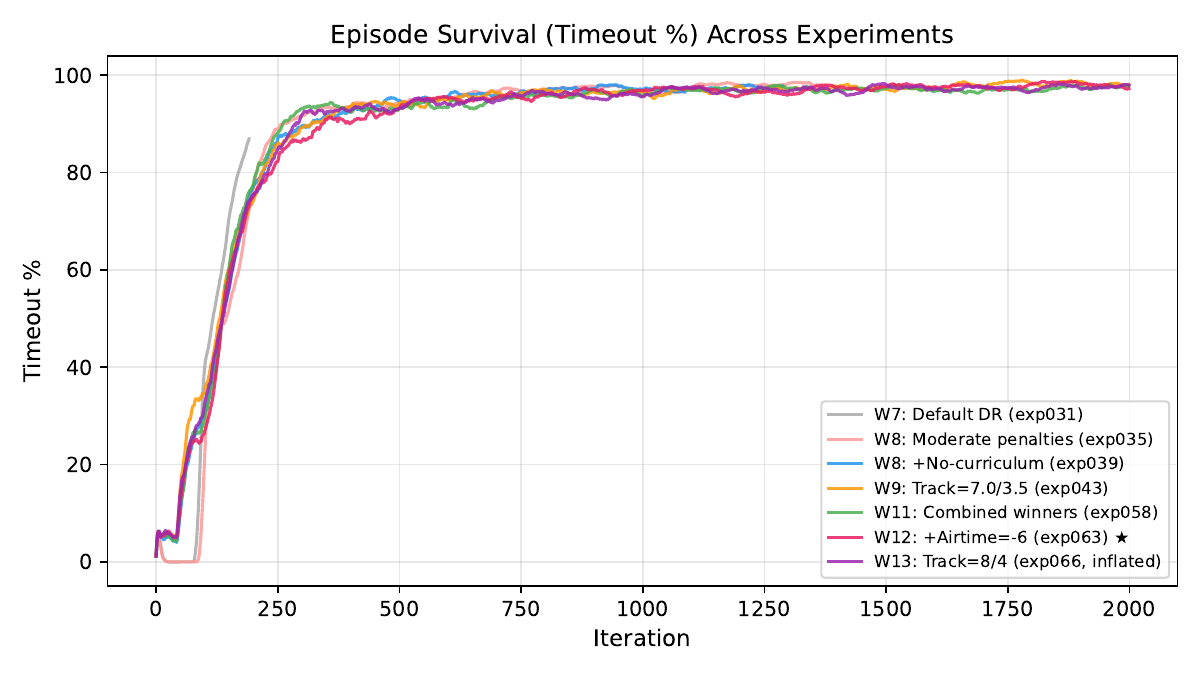}
\caption{Episode survival rate (timeout \%) across key experiments. Higher values indicate fewer premature terminations due to falls. The best configurations sustain $>$95\% timeout rates, indicating stable locomotion throughout most episodes.}
\label{fig:timeout}
\end{figure}

Fig.~\ref{fig:timeout} shows the episode survival rate across the same key experiments. Higher timeout percentages indicate that robots completed full episodes without falling. The best configurations (exp043, exp058, exp063) sustain timeout rates above 97\%, showing that the refined reward recipes produced not only higher reward but also more stable locomotion.

\subsection{Reward Scale Normalization}\label{sec:normalization}
An important methodological note: comparing total mean reward across experiments with different reward weight configurations can be misleading. For example, Wave 9 experiments with tracking weights of 7.0/3.5 reported total reward $\sim$155 compared to $\sim$116 for Wave 8 runs with weights 5.0/2.5. However, when the per-component tracking reward is normalized by its weight (i.e., dividing the reported component value by its coefficient), the underlying tracking quality is similar ($\sim$0.88) across configurations. The scale-independent velocity error metric confirms this: experiments with moderate tracking weights (5.0/2.5) actually achieved the lowest planar velocity errors ($\sim$0.29), while experiments with stronger tracking weights (7.0/3.5) converged to slightly higher errors ($\sim$0.32). This finding illustrates a general lesson: the agent's ability to interpret reward decompositions and recognize when total reward increases reflect scale changes rather than behavioral improvements is important for avoiding false conclusions in autonomous research.

\begin{figure}[t]
\centering
\includegraphics[width=\columnwidth]{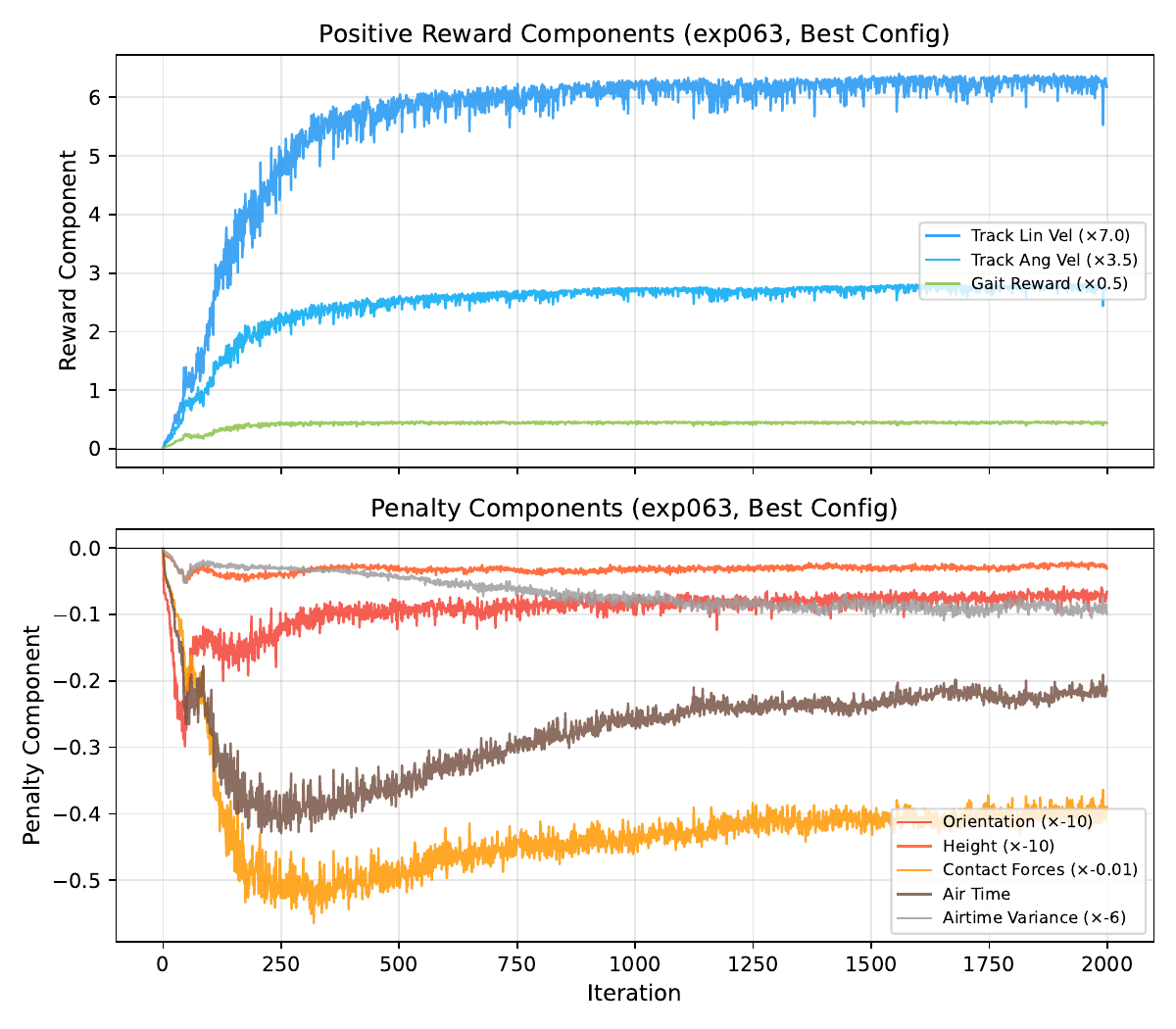}
\caption{Reward decomposition for exp063 (best configuration). Top: positive components (tracking, gait). Bottom: penalty components. At convergence, tracking rewards dominate while stability penalties are near zero, indicating the policy learned stable locomotion.}
\label{fig:decomp}
\end{figure}

Fig.~\ref{fig:decomp} shows the full reward decomposition for the best run (exp063). The top panel reveals that tracking rewards dominate the total signal, with linear velocity tracking ($\times 7.0$) contributing roughly 6 units and angular velocity tracking ($\times 3.5$) contributing roughly 2.7 units at convergence. The bottom panel shows that penalty magnitudes are comparatively small: orientation and height penalties converge to near zero (indicating the policy learned to maintain stable posture), while contact forces remain the largest penalty at approximately $-0.4$. This decomposition confirms that the final policy achieved a healthy reward balance where tracking incentives clearly outweigh penalties, unlike early waves where penalties nearly canceled tracking rewards.

\subsection{Agent-Driven Versus Typical Human-Led Iteration}
This archive does not include a controlled human-versus-agent ablation, so any comparison must remain qualitative. Still, two differences are visible. First, the agent handled many low-level iteration costs that a human researcher often defers or batches, such as repeated code inspection, log comparison, run triage, and reference-to-code transfer. Second, the agent was willing to execute narrow diagnostic experiments that only answered one question, such as whether fewer environments changed the deadlock behavior. In a human-led workflow those checks might be delayed in favor of more intuitive reward tuning.

That said, the human still set the overall agenda. The record is better described as autonomous execution within human-defined research directions than as full research independence.

\section{Discussion}
The case study suggests that agent-driven research becomes more useful when the bottleneck is not purely mathematical. Much of the progress here came from mixed reasoning over code, logs, runtime errors, reward terms, and simulator behavior. The agent's value was not only in proposing reward changes, but also in turning vague failures into narrower engineering hypotheses.

Several limitations are equally clear. The human provided the top-level prompts and decided what broad topics mattered. The archive reflects a single project rather than a benchmark over many robotics RL tasks. Some evidence is log-based rather than derived from standardized evaluation episodes. Finally, the process depended on an agentic environment with shell access, file editing, and experiment control, so the results should not be read as evidence about standalone chat interfaces.

Compared with prior autonomous research systems, this work sits between full automation and tool-augmented collaboration. AutoResearch emphasizes a nearly closed loop on a compact ML setup \cite{autoresearch}. The AI Scientist \cite{aiscientist, aiscientistv2} operates in well-scoped ML domains with standardized evaluation. Our case study is messier and more realistic for robotics: multiple GPUs, simulator bootstrapping, terrain deadlocks, reward ports from external code, and partial human steering. Unlike Eureka \cite{eureka}, where the agent generates reward functions from scratch, our agent identified existing reward terms from reference code and tuned their weights through iterative experimentation. That difference is not a weakness of the comparison. It is the point of it.

\section{Conclusion}
This paper presented the experiment archive as a case study in agent-driven autonomous reinforcement learning research. In this setting, a human supplied high-level goals and an agent executed most of the detailed loop across fourteen waves and more than 70 experiments. The clearest outcomes were the diagnosis of a terrain-linked PhysX deadlock, the transfer of four reward terms from an openly available reference implementation, the decision to de-emphasize HIM after repeated terrain=0.0 results, and the eventual improvement from early reward levels around 7 to a best logged Wave 12 result with velocity error 0.263 and 97\% timeout over 2000 complete iterations in exp063, independently reproduced five times across three GPUs. The agent discovered that reducing gait enforcement, strengthening stability penalties, and increasing airtime variance regularization improved velocity tracking---non-obvious findings that emerged from systematic single-variable experimentation, then confirmed by combining winning factors and reproducing results across GPUs. Two additional waves (13--14) systematically explored the remaining configuration space---gait weight, penalty magnitudes, tracking weights, airtime variance, contact forces, and auxiliary reward terms---without finding any improvement, confirming convergence.

The contribution is documentary rather than universal. In this case, an agent materially advanced a robotics RL project by combining code editing, experiment design, log analysis, and runtime triage inside an agentic environment. That does not show that such agents can replace human researchers in general. It does show that they can carry a substantial share of the iterative RL research workload in a realistic, failure-prone setting.

\bibliographystyle{IEEEtran}
\bibliography{references}

\end{document}